\documentclass[conference]{IEEEtran}
\IEEEoverridecommandlockouts
\usepackage{cite}
\usepackage{amsmath,amssymb,amsfonts}
\usepackage{algorithmic}
\usepackage{graphicx}
\usepackage{textcomp}
\usepackage{xcolor}
\usepackage{caption}
\usepackage{hyperref}
\usepackage[labelformat=simple]{subcaption}
\usepackage[linesnumbered,ruled]{algorithm2e}

\newtheorem{definition}{Definition}
\newtheorem{proposition}{Proposition}

\newtheorem{formulation}{Problem Formulation}

\DeclareMathOperator*{\argmin}{arg\,min}

\def\BibTeX{{\rm B\kern-.05em{\sc i\kern-.025em b}\kern-.08em
    T\kern-.1667em\lower.7ex\hbox{E}\kern-.125emX}}
\begin{document}

\title{Striking a Balance in Fairness for Dynamic Systems Through Reinforcement Learning}

\author{\IEEEauthorblockN{Yaowei Hu, Jacob Lear, Lu Zhang}
\IEEEauthorblockA{\textit{Department of Electrical Engineering and Computer Science} \\
\textit{University of Arkansas}\\
Fayetteville, Arkansas, USA \\
\{yaoweihu, jdlear, lz006\}@uark.edu}
}

\maketitle

\begin{abstract}
While significant advancements have been made in the field of fair machine learning, the majority of studies focus on scenarios where the decision model operates on a static population. In this paper, we study 
fairness in dynamic systems where sequential decisions are made.
Each decision may shift the underlying distribution of features or user behavior. We model the dynamic system through a Markov Decision Process (MDP). By acknowledging that traditional fairness notions and long-term fairness are distinct requirements that may not necessarily align with one another, we propose an algorithmic framework to integrate various fairness considerations with reinforcement learning using both pre-processing and in-processing approaches. Three case studies show that our method can strike a balance between traditional fairness notions, long-term fairness, and utility.
\end{abstract}


\section{Introduction}
Machine learning algorithms are increasingly being widely used in high-stakes decision-making applications, such as college admissions \cite{johnson2016impartial, baker2021algorithmic}, bank loans \cite{Zhang2017AchievingNI,lee2021algorithmic}, employment \cite{schumann2020we}, recidivism risk assessment \cite{berk2021fairness}, etc. However, our society is rife with systemic inequalities, and real-world data is often influenced by social and historical contexts which may contain historical biases and discrimination based on gender, race, and other factors. As a result, algorithmic fairness in machine learning has received increasing attention. Ensuring that machine learning models trained on biased data do not make discriminatory decisions against vulnerable groups is a key consideration for deploying these algorithmic decision-making systems \cite{li2022fairness}. In the literature, various notions and metrics have been proposed to measure the concept of fairness in machine learning. Diverse algorithms have been proposed to address the fairness concerns. 
For surveys on general fair machine learning, please refer to \cite{tang2022and,pessach2022review,alves2023survey}.

Despite the progress in fair machine learning, most of the studies focus on scenarios in which the decision model makes decisions over a static population. These works typically employ fair machine learning algorithms to achieve fairness for a batch of data, but do not consider how these decisions may affect fairness over an extended period of time. Nevertheless, real-world decision-making systems usually operate in a dynamic manner that involves making sequential decisions. Each decision may shift the underlying distribution of features or user behavior, which in turn affects the subsequent decisions. For example, loan decisions have an impact on an individual's credit score, while the allocation of police forces can influence the crime rate in a particular location.
As a result, long-term fairness has been proposed as a fairness notion that concerns how fairness can be achieved and maintained over a time horizon, rather than a single time step, by taking into consideration the system dynamics and feedback loops.


How to achieve long-term fairness has been explored in previous studies through explicit modeling of dynamics and feedback loops. One branch of research is to investigate the impact of current decisions on a target population in application-specific scenarios by leveraging various analytical frameworks, such as the one-step feedback \cite{DBLP:conf/icml/LiuDRSH18} or Pólya urn model \cite{ensign2018runaway}. There are analytical results that show that 
simply enforcing traditional fairness notions at each static decision point may produce adverse effects on disadvantaged groups in the long run \cite{liu2018delayed,zhang2020fair}.
Long-term fairness has also been studied in the context of reinforcement learning (RL) where the system dynamics and feedback loops between decisions and the population are formulated through a Markov Decision Process (MDP) \cite{jabbari2017fairness}. Following this line of research, \cite{d2020fairness,atwood2019fair} establish simulation environments for studying long-term fairness in RL, based on which RL algorithms have been used to learn a decision-making policy that aims to optimize both policy utility and fairness as long-term objectives \cite{yu2022policy, wen2021algorithms}.
However, one limitation of existing methods is that they usually formulate long-term fairness as the difference between the average or instantaneous rewards received by different demographic groups similar to static fairness notions, but do not take into account the inherent difference between traditional fairness notions and long-term fairness requirements.



In this paper, we consider long-term fairness as a requirement on the states rather than the rewards following the definitions in some well-accepted research \cite{liu2018delayed,hu2018short,zhang2020fair}. We acknowledge that traditional fairness notions and long-term fairness are distinct requirements that may not necessarily align with one another. Traditional fairness considers the equity of the outcomes or performance of the decision model at a single decision point. It is referred to as short-term fairness later in this paper for a clear representation. Long-term fairness, on the other hand, refers to a long-term state in which equity is systematically satisfied. Such a state may be achieved by gradually reducing the gap between the qualification distributions of different groups. As a result, imposing short-term fairness constraints may not necessarily lead to long-term fairness even if the constraints are incorporated into a long-term objective. For example, suppose a bank uses different thresholds for making loan decisions for the advantaged and disadvantaged groups in order to ensure fair outcomes. However, this approach may not help narrow
the gap between the credit score distributions of the two groups.


To address the above issue, we develop an algorithmic framework that promotes both short-term and long-term fairness simultaneously. Similar to prior works, we utilize the MDP framework to leverage its power in optimizing long-term objectives. By recognizing the distinct requirements of short-term and long-term fairness, we incorporate them into the RL algorithm using different approaches. Since the concept of long-term fairness is aligned with the principle of the MDP framework, we employ an in-processing approach to deal with this constraint. We adopt the 1-Wasserstein distance as the metric of the distribution gap and theoretically show that minimizing the distance can lead to a long-term fair state. On the other hand, we adopt a model-agnostic pre-processing approach to deal with short-term fairness to ensure that it is enforced throughout the sequential decision-making process. We extend a classic pre-processing approach called massaging \cite{kamiran2009classifying} to the RL setting by integrating it with the policy optimization algorithm. Finally, we show the exact implementation of our algorithmic framework using three case studies, where the experimental results demonstrate that our method is capable of striking a desired balance between short-term fairness, long-term fairness, and the utility of the sequential decision system.

We summarize our contributions as below:
\begin{itemize}
    \item 
    We propose to achieve systematic equity in sequential decision-making by considering short-term and long-term fairness as distinct fairness requirements. 
    \item 
    We develop an efficient and flexible algorithmic framework that integrates short-term and long-term fairness with the MDP framework as distinct constraints. 
    \item Three case studies within simulation environments are used to prove the effectiveness of our method by evaluating the performance of our method and comparing it with the state-of-the-art baselines. 
\end{itemize}

\section{Related Work}
Extensive research has been conducted on achieving fairness in machine learning within static systems, but the exploration of fairness in dynamic systems has remained relatively limited. It has been first studied in a compound decision-making process called a pipeline \cite{bower2017fair,dwork2018fairness} where individuals may drop out at any stage and classification in subsequent stages depends on the remaining cohort of individuals.
Then, more complex dynamic systems are considered where the sequential decisions have the potential to reshape the distribution of the underlying population. Due to the difficulty of modeling the system dynamics, researchers first focused on specific applications. For example, the authors in \cite{hu2018short} propose a short-term intervention for long-term fairness in the labor market by constructing a dynamic reputational model and adopting a dual labor market composed of a Temporary Labor Market and a Permanent Labor Market. 
The research in \cite{liu2018delayed} studies the delayed impact of fair machine learning in lending scenarios. It has been realized that, traditional fairness notions cannot guarantee to promote fairness as a delayed impact, even in a one-step feedback model. Researchers in \cite{zhang2020fair} also verify the inadequacy of traditional fairness notions in achieving fairness in the long run. It has also been proposed to model the dynamic system using causal graphs. \cite{creager2020causal} introduces causal directed acyclic graphs (DAGs) as a framework for studying fairness in dynamical systems. 
The research in \cite{hu2020fair,hu2022achieving} proposes to treat model deployment as soft interventions and infer post-intervention distributions in measuring fairness.

As Markov Decision Process (MDP) is a prevalent framework used to model sequential decision-making, it has also been proposed to study fairness in the context of reinforcement learning (RL). The research in \cite{jabbari2017fairness} initiates the study of fairness in RL and proposes a fairness constraint that requires an algorithm to never prefer one action over another if the long-term reward of choosing the latter action is higher. However, fairness is defined only based on the reward of each action without considering the demographic information. Along this direction, the authors in \cite{ge2021towards} study long-term fairness and formulate a Constrained Markov Decision Process (CMDP), but fairness is defined only within the context of recommendation systems.
The authors in \cite{wen2021algorithms} propose two algorithms for learning policies to satisfy fairness constraints that are defined on the average reward for individuals in different groups.
In particular, \cite{d2020fairness,atwood2019fair} advocate for the use of simulation studies to understand the long-term behaviors of deployed ML-based decision systems and their potential consequences. The authors explore toy examples of dynamical systems including bank loans, college admissions, allocation of attention, and epidemic control. An extensible open-source software framework for implementing fairness-focused simulation studies is provided. Based on that, \cite{yu2022policy} proposes to impose fairness requirements in policy optimization by regularizing the advantage evaluation of different actions. The proposed methods make it easy to impose fairness constraints without rewarding engineering or sacrificing training efficiency. However, the authors directly apply traditional fairness notions without considering the different requirements of traditional and long-term fairness. Different from prior works, we consider long-term fairness as a requirement on the states rather than on the rewards and acknowledge traditional and long-term fairness as distinct fairness requirements. We develop an algorithmic framework that promotes both requirements simultaneously.

\section{Preliminaries}
This section introduces the background and preliminaries of fair machine learning, reinforcement learning (RL), and proximal policy optimization (PPO), a prevalent policy optimization algorithm in RL.


\subsection{Fair Machine Learning}
The issue of fairness has become one of the most popular topics in machine learning in recent years. To measure fairness in algorithmic decision-making, a large number of fairness notions have been proposed in the literature. Typical examples include \textit{demographic parity (DP)} and \textit{equal opportunity (EO)}. DP aims to ensure that different demographic groups are represented proportionally in the outcomes of a decision model. 
EO, on the other hand, refers to the principle of treating individuals or groups fairly by ensuring equal error rates or predictive performance across different demographic subgroups. Then, to address the fairness issues, bias mitigation algorithms are proposed mainly from three perspectives: pre-processing, in-processing, and post-processing. Pre-processing approaches focus on eliminating bias from the training data (e.g., \cite{kamiran2012data,feldman2015certifying,xu2018fairgan}), in-processing approaches aim to avoid introducing bias in model training by proposing new model structures or loss functions (e.g., \cite{quadrianto2017recycling,kim2019preference,zafar2017parity}), and post-processing approaches modify predicted outcomes to resolve fairness issues (e.g., \cite{Hardt2016EqualityOO,corbett2017algorithmic,menon2018cost}).



\subsection{Reinforcement Learning}
Reinforcement learning consists of two interactive components, an agent and an environment, which interact with each other over time. This interaction process is modeled as a Markov decision process (MDP) \cite{sutton2018reinforcement}. An MDP is denoted by a tuple $M = (\mathcal{S}, \mathcal{A}, P, R, \rho_0, \gamma)$, where $S \in \mathcal{S}$ is a set of states, $A \in \mathcal{A}$ is a set of actions, $P: \mathcal{S} \times \mathcal{A} \times \mathcal{S} \to [0, 1]$ is a transition function that represents the probability of next state given the current state and the action, $R: \mathcal{S} \to \mathbb{R}$ is a reward function, $\rho_0: \mathcal{S} \to [0, 1]$ is an initial state distribution, and $\gamma \in [0, 1]$ is a discount factor. At each time step $t$, the agent observes a state $s_t \in S$ from the environment and takes an action $a_t \in A$ following a policy $\pi: \mathcal{S} \to \mathcal{A}$ based on the current state. Then the agent observes a new state $s_{t+1} \in S$ and a reward $r_t \in R$ generated by the environment with the transition probability $P(s_{t+1}|s_t, a_t)$ and the reward function $R(s_t)$ in the next time step $t+1$. The goal of RL is to learn a optimal policy $\pi_\theta$ which maximizes the expected discounted cumulative rewards, defined as below.

$$J(\theta) = \mathbb{E}_{\tau \sim \pi_\theta}\left[\sum_{t=0}^\infty \gamma^tR(s_t)\right],$$ 
where $\tau = (s_0, a_0, s_1, a_1, ...)$ is a trajectory and $\tau \sim \pi_\theta$ means that a trajectory $\tau$ is sampled from the policy $\pi_\theta$ following $s_0 \sim \rho_0(s_0), a_t \sim \pi(a_t|s_t), s_{t+1} \sim P(s_{t+1}|s_t, a_t)$.


To address the RL problem, there are several concepts that are often involved in RL algorithms.
Let $R(\tau)$ denote the discounted cumulative rewards of a trajectory $\tau$. The state value function $V$ and the state-action value function $Q$ are given by $V(s_t) = \mathbb{E}_{\tau \sim \pi}[R(\tau)|s_t = s]$ and $Q(s_t, a_t) = \mathbb{E}_{\tau \sim \pi}[R(\tau)|s_t = s, a_t = a]$, which evaluate how good a state or a pair of state and action is. The advantage function is the difference between $Q(s_t, a_t)$ and $V(s_t)$, i.e., $A(s_t, a_t) = Q(s_t, a_t) - V(s_t)$, and it can be considered as the advantage of taking a given action over following the policy \cite{schulman2015high}.

\subsection{Proximal Policy Optimization}
Policy optimization methods \cite{sutton1999policy} are a type of reinforcement learning algorithms that improve policies directly by estimating policy gradients and optimizing with stochastic gradient ascent. The most commonly used form of gradient estimator is given by
\begin{equation}
\label{eq:gd}
\nabla J(\theta) = \mathbb{E}_{(s_t,a_t)\sim \pi_{\theta}}[A(s_t,a_t) \nabla_\theta \log \pi_\theta(a_t|s_t)]
\end{equation}
where 
the expectation is estimated over a batch of samples.

Proximal policy optimization (PPO) \cite{schulman2017proximal} is a state-of-the-art policy optimization algorithm stemming from the trust region policy optimization (TRPO) algorithm \cite{schulman2015trust}. It maximizes a clipped surrogate function to prevent the gradient update dramatically, as follows
\begin{equation}
\label{eq:ppo}
J^{\mathit{CLIP}}(\theta) = \mathbb{\hat{E}}\left[ \min(r_t(\theta)A_t, \text{clip}(r_t(\theta), 1-\epsilon, 1+\epsilon)A_t)  \right]    
\end{equation}
where $A_t$ is short for $A(s_t,a_t)$, $r_t(\theta)$ denotes the probability ratio $\frac{\pi_\theta(a_t|s_t)}{\pi_{\theta_{old}}(a_t|s_t)}$, and $\epsilon$ is a hyperparameter. To compute variance-reduced advantage function, a neural network is used to estimate the state value function $V(s_t)$ with the squared-error loss
\begin{equation}
\label{eq:loss2}
    L^V(\theta) = \mathbb{E}[(V_\theta(s_t) - R(\tau))^2]
\end{equation}
The clipped surrogate function restricts the magnitude of the gradient update, which not only makes the algorithm more stable, but also allows for multiple updates using a batch of samples, improving the data efficiency. 

In this paper, we adopt PPO as the RL algorithm, but our method can be applied to any policy optimization algorithms.


\section{Problem Formulation}
To develop an RL algorithm for achieving long-term fairness, we start by defining fairness notions in the context of sequential decision-making, presenting the problem formulation, and establishing the setting of the fair RL learning problem.



\subsection{Fairness Definition for Sequential Decision Making}
As stated in the last section, traditional fairness notions are usually concerned about whether the machine learning model produces the same outcome or performance across different groups in a static population.
In the context of sequential decision-making, we refer to this type of notions as short-term fairness notions, which are defined based on a cohort of individuals who participate in the decision-making system over a specific period. 
\begin{definition}[Short-term Fairness]
In a sequential decision-making system, short-term fairness is defined as the equal outcome or performance of the decision model/policy over a participating cohort.
\end{definition}

It is worth emphasizing that, first, short-term fairness notions may be enforced by laws and regulations in some domains, such as the U.S. Equal Employment Opportunity Commission \cite{barocas2017fairness} that prohibits employment discrimination. So, it is essential to enforce short-term fairness throughout the sequential decision-making process including both training and evaluation. Second, different short-term fairness notions may conflict and not be achieved simultaneously if historical and/or systemic biases exist \cite{alves2023survey}. For example, achieving demographic parity may require preferential treatment to account for historical disadvantages, which could potentially impact equal opportunity. Thus, we adopt a single short-term fairness notion only in our algorithm.

On the other hand, long-term fairness has been proposed to account for fairness and equity of the sequential decision-making system in the long run \cite{liu2018delayed}. The general goal of long-term fairness is to reach a state where the historical disadvantages are rectified and systemic biases are removed. Since the trade-off between short-term and long-term fairness is due to historical disadvantages and systemic biases, we presume that long-term fairness also implies a state in which it becomes easier to simultaneously satisfy different short-term fairness notions. 

{\bf\noindent How to quantify long-term fairness}? In the literature, features are regarded as indicators or metrics that assess the qualification or competency levels of individuals.
Then, long-term fairness is often formulated by measuring the gap in feature distributions between different groups. 
For example, in \cite{liu2018delayed}, the difference in feature distribution of the disadvantaged group between the starting time step and the ending time step, i.e., $\bigtriangleup = \mathbb{E}[ x_{t=t^*} | c^- ] - \mathbb{E}[ x_{t=0} | c^- ]$, 
is defined as the measure of long-term fairness. 
It is called long-term improvement if $\bigtriangleup > 0$, stagnation if $\bigtriangleup = 0$, and active harm if $\bigtriangleup < 0$. In \cite{hu2018short}, long-term fairness is defined as the parity in the feature distribution between the advantaged and disadvantaged group when the dynamic system is at equilibrium, i.e., $\bigtriangleup = |\mathbb{E}[ x_{t=t^*} | c^+ ] - \mathbb{E}[ x_{t=t^*} | c^- ]|$.
In this paper, we adopt a similar philosophy to \cite{hu2018short} and define long-term fairness as follows.
\begin{definition}[Long-term Fairness]
Long-term fairness is defined as the equal feature distributions of different groups at a long-term state of the sequential decision-making system.
\end{definition}

After discussing the two types of fairness we consider, we define our problem formulation as follows.
\begin{formulation}
Consider a sequential decision-making context. A policy for making the decision is learned through an iterative process of interaction with the environment. Our goal is to learn a fair policy such that: (1) short-term fairness is guaranteed throughout both the training and evaluation processes, and (2) a long-term fair state is reached at the end of the evaluation process.
\end{formulation}

\subsection{Problem Setting for Fair RL}
To formulate the problem of long-term fair sequential decision-making,
we consider a finite-horizon RL problem and leverage the MDP framework. Specifically, in our context states $\mathcal{S} = \mathcal{C} \times \mathcal{X}$ where $\mathcal{C}$ is the domain of the sensitive feature and $\mathcal{X} \in \mathbb{R}^m$ represents the domain of the profile features. When $\mathcal{C}$ is a binary domain, we use $\{c^+, c^-\}$ to represent the advantaged and disadvantaged groups respectively. Let $\mathcal{A}$ denote the action space. If $\mathcal{A}$ is a binary domain, we use $\{1, 0\}$ to represent the positive and negative actions. Our goal is to learn a stochastic policy $\pi_\theta: \mathcal{S} \to \mathcal{A}$ which maximizes the agent's cumulative reward while satisfying certain fairness criteria.
Note that we generally allow the sensitive attributes to be involved in the policy input and will explicitly adopt constraints to enforce fairness requirements.


We may use the bank loan system as an example to illustrate this problem setting in a real-world scenario. In this example, the bank is treated as the agent, and the population of the applicants is treated as the environment. Assume that the sensitive attribute $C$ is the race of the applicant. As an RL process, at each time step $t$, an individual from a certain race group is sampled and applies for a loan from the bank, whose state $s_t$ is given by race $c_t$ and personal profile $x_t$. Then, the bank runs a policy function $\pi_\theta$ to decide whether to approve the loan according to probability distribution $\pi_{\theta}(a_t|s_t)$. If the loan is approved, depending on whether the individual repays the loan, both the bank's profit and the feature distribution of the population will be affected which determines the new state and the reward. Finally, the goal of the bank is to learn from interactions a lending policy that maximizes its own profit while satisfying fairness.



\section{Algorithm}


In this section, we develop a flexible and effective fair RL algorithm that integrates the pre-processing and in-processing approaches to promote short-term and long-term fairness simultaneously. We assume separate training and evaluation processes where the policy is updated according to the interaction experience with the environment during the training process, and it is further evaluated in a separate environment with the same dynamics after the training completes. However, our method is readily applied to the setting where the evaluation is conducted during the training. In the following, we first provide an overview of the algorithm.


\subsection{Overview}
When developing the fair RL algorithm, it is critical to consider the different requirements for long-term and short-term fairness. Long-term fairness is a state that represents the maximization of the equity of the system in the long run. As it is aligned with the principle of the MDP framework that maximizes the expected total reward over time, we adopt an in-processing approach that regularizes the reward to incorporate the long-term fairness objective so that the RL algorithm can be aware of the long-term fairness status in training. Specifically, we regularize the advantage function of a policy optimization algorithm as:
\begin{equation}\label{eq:adv}
    A{^\lambda}(s_t, a_t) = A(s_t, a_t) + \lambda R(s_t),
\end{equation}
where $R(s_t)$ is the regularization that reflects the long-term fairness requirements 
and $\lambda$ is a hyperparameter that controls the degree of regularization.

On the other hand, short-term fairness is an instantaneous constraint that may be enforced by laws or regulations at every step of the decision-making. 
Merely incorporating short-term fairness through regularizing the advantage function may not be sufficient to guarantee short-term fairness throughout the entire training process. Thus, we choose to adopt a pre-processing approach to address this issue. Inspired by the classic pre-processing approach named massaging \cite{kamiran2009classifying}, we propose a method called \textit{action massaging} that selectively alters unfair actions produced by the policy network to fair ones. Specifically, after an action $a_t$ is sampled from the policy network $\pi_\theta$ at time step $t$, we employ a functional mapping $a'_{t}=m(s_t, a_t)$ where $a'_t$ may or may not be equal to $a_t$ to generate the trajectory. By using this altered trajectory to perform policy optimization, the policy gradient becomes
\begin{equation}
\label{eq:gd'}
\nabla J(\theta) = \mathbb{E}_{(s_t,a_t)\sim \pi_{\theta}}[A^{\lambda}(s_t,a'_t) \nabla_\theta \log \pi_\theta(a'_t|s_t)],
\end{equation}
which shows that the policy gradient is computed based on the trajectory formed by $a'_t$.
The rationale of the action massaging is to perform fair actions when the policy network generates biased ones and also encourage the policy to generate fair and high-reward actions. Note that this approach differs from the off-policy RL learning algorithm which optimizes the current policy network based on the trajectories generated by a different policy, and hence our approach does not require importance sampling to correct for the bias.

Next, we describe the above pre-processing and in-processing components in detail.

\subsection{Action Massaging for Short-term Fairness}

The action massaging altered actions according to a pre-defined short-term fairness criterion that is to meet legal and regulatory requirements for decision-making and prevent discrimination against certain groups. As mentioned earlier, we consider group fairness notions such as DP or EO in our work. At each time step $t$, short-term fairness relies on the current state and action as well as the past states and actions of the system. To facilitate computation, we adopt a sliding window $w$ so that all the states and actions between time step $t-w$ and $t$ form a participating cohort that will be used to measure short-term fairness. We denote this measure as $\bigtriangleup_s(s_t, a_t)$. Then, the action massaging $m(s_t, a_t)$ alters the action $a_t$ to $a'_t$ that minimizes $\bigtriangleup_s(s_t, a_t)$ to improve short-term fairness.

When designing the mapping $m(s_t, a_t)$, one principle is that the modifications should not significantly damage the utility of the policy. In our method, we treat $\pi_\theta(a_t|s_t)$ as the confidence level for selecting action $a_t$, and introduce a constraint that limits the difference between the confidence of the original action $a_t$ and the altered action $a'_t$. The action massaging only alters the current action to a different action when the above difference is smaller than a predefined threshold. When multiple actions satisfy the constraint, the action massaging chooses the one that leads to best short-term fairness. As a result, the action massaging is formulated as follows:
\begin{align}\begin{aligned}
    \label{eq:am}
    m(s_t,a_t) = & \argmin_{a'_t \in \mathcal{A}}  \bigtriangleup_s(s_t, a'_t) \\
    & \mathrm{s.t.} \quad | \pi_\theta(a_t|s_t)-\pi_\theta(a'_t|s_t) | < \tau.
\end{aligned}\end{align}

The constraint in Eq.~\eqref{eq:am} is an important factor that reflects the trade-off between short-term fairness and utility. It aids in enhancing short-term fairness while minimizing the impact on utility. On one hand, the constraint reduces the number of modifications made by the action massaging, as a large number of modifications will cause instability and deviation of training. On the other hand, the constraint also restricts the modifications to be carried out when current actions have low confidence and hence leads to a smaller reduction in utility. For example, in a special case of binary actions (e.g., the decision of bank loan), the constraint will restrict the modifications to actions with confidence close to 0.5. The exact implementation of the action massaging is task-specific and varies with applications, as will be shown in the case studies in the next section.


\subsection{Advantage Regularization for Long-term Fairness}
The action massaging for short-term fairness is not enough to achieve long-term fairness as the objective of short-term fairness may not exactly align with the objective of long-term fairness. As mentioned above, we leverage the MDP's capacity to maximize long-term returns as a means to attain long-term fairness. To quantify long-term fairness, in our work we employ the 1-Wasserstein distance between the distributions of different groups as the long-term fairness metric. The reason is presented in the following proposition.

\begin{proposition}
Denote by $d$ the 1-Wasserstein distance between the feature distributions of different groups, i.e., 
$d = W(P(x|c^+),P(x|c^-))$.
For any decision model $h: \mathcal{X} \mapsto \mathcal{A}$ that is Lipschitz continuous, its DP is bounded by $l_h \cdot d$ where $l_h$ is the Lipschitz constant of $h$. If we assume that the true label is given by another decision model $g: \mathcal{X} \mapsto \mathcal{A}$ that is Lipschitz continuous and also satisfies DP, then the EO of $h$ is bounded by $(l_h+l_g)/P(y)\cdot w$ where $l_g$ is the Lipschitz constant of $g$.
\end{proposition}

\begin{IEEEproof}
According to the definition of DP, we have
\begin{equation*}
    \mathrm{DP}(h) = |\mathbb{E}[h(x)|c^+] - \mathbb{E}(h(x)|c^-)|.
\end{equation*}
Due to the Kantorovich–Rubinstein duality \cite{villani2021topics}, it is straightforward that
\begin{equation*}
\begin{split}
    \mathrm{DP}(h) & \leq \sup_{\lVert h \rVert \leq l_h} \left[ \mathbb{E}_{x\sim P(x|c^+)}[h(x)] - \mathbb{E}_{x\sim P(x|c^-)}[h(x)] \right] \\
    & = l_h \cdot W(P(x|c^+),P(x|c^-)) = l_h \cdot d.
\end{split}
\end{equation*}
On the other hand, we have
\begin{equation*}
    \mathrm{EO}(h) = |\mathbb{E}[h(x)|a=1, c^+] - \mathbb{E}(h(x)|a=1, c^-)|.
\end{equation*}
Due to the assumption that the true label is given by $g$ and $g$ satisfies DP, it follows that
\begin{equation*}
\begin{split}
    & \mathbb{E}[h(x)|a, c] = \int_{x} h(x)P(x|a,c) dx \\
    & = \int_{x} h(x)P(x|c)\frac{P(y|x,c)}{P(y|c)} dx = \int_{x} h(x)P(x|c)\frac{g(x)}{P(y)} dx \\
    & = \frac{1}{P(y)}\mathbb{E}_{x\sim P(x|c)}[h(x)g(x)].
\end{split}
\end{equation*}
In addition, define $f(x)=h(x)g(x)$ and denote the Lipschitz constant of $f$ as $l_f$. It is easy to show that $l_f \leq l_h \cdot \sup_x |h(x)| + l_g \cdot \sup_x |g(x)|$. Since $h(x)\leq 1$ and $g(x)\leq 1$, we have $l_f\leq l_h+l_g$. As a result, we have
\begin{equation*}
    EO(h) \leq \frac{l_h+l_g}{P(y)} W(P(x|c^+),P(x|c^-)) = \frac{l_h+l_g}{P(y)} \cdot d.
\end{equation*}

\end{IEEEproof}

The above proposition shows that, by approaching a long-term state where the 1-Wasserstein distance between the feature distributions of different groups is minimized, we can mitigate at that state both the DP and EO of any decision model that is Lipschitz continuous. This implies that a long-term fair state has been reached.


Denote the long-term fairness measure computed at time step $t$ as $\bigtriangleup_l(s_t)$. Similar to short-term fairness, we adopt a sliding window to form a participating cohort for estimating the feature distributions. Then, we incorporate $\bigtriangleup_l(s_t)$ into the advantage function as the regularization. However, rather than directly adding $\bigtriangleup_l(s_t)$ to the advantage function, we further consider the trade-off between short-term fairness and long-term fairness when the two objectives are not aligned. Specifically, we promote the advantage when both short-term and long-term fairness can be improved while demoting the advantage when both short-term and long-term fairness is damaged. When there is a conflict between short-term and long-term fairness, we keep the current advantage unchanged. The regularization term is defined as follows:
\begin{equation}
\label{eq:reg}
\begin{split}
R(s_t) = 
&\begin{cases}
\min(0, \bigtriangleup_l(s_t) - \bigtriangleup_l(s_{t+1})) & \bigtriangleup_s(s_t, a_t) > \delta\\
\max(
0, \bigtriangleup_l(s_t) - \bigtriangleup_l(s_{t+1})) & \bigtriangleup_s(s_t, a_t) \le \delta\\
\end{cases}    
\end{split}
\end{equation}


As can be seen, the first term is active when short-term fairness $\bigtriangleup_s(s_t, a_t)$ is larger than the threshold $\delta$. Then, this term will penalize the original advantage when the long-term fairness measure does not decrease at the next time step $t+1$ compared to the current time step $t$. The second term is active when short-term fairness $\bigtriangleup_s(s_t, a_t)$ is less or equal to $\delta$. In this case, we reward the advantage function when the long-term fairness measure reduces. This approach allows for a continuous improvement of long-term fairness throughout the entire sequence, rather than a sudden change at a specific point, while it remains simple and effective. 


Combining the above two methods for fairness, we present the pseudocode of our algorithm in Algorithm \ref{algo:3}, which is referred to as F-PPO.

\IncMargin{1em}
\begin{algorithm}[t]\small
\SetKwInOut{Input}{Input}
\SetKwInOut{Output}{Output}
\SetKwRepeat{Repeat}{repeat}{until}
\SetKw{Return}{return}
	\caption{\textbf{Fair Proximal Policy Optimization (F-PPO)}}
    \label{algo:3} 
	\BlankLine 
	Initialize policy network $\pi_\theta$ and value function network $v_\phi$\;
	
	\For {k = \rm{0, 1, 2, ...}} {
	    Collect trajectories $\mathcal{D}_k$ from policy $\pi_\theta$ where actions $a_t$ are sampled from $\pi_\theta(a_t|s_t)$\;

        Compute $\bigtriangleup_s(s_t, a_t)$ and apply action massaging according to Eq.~\eqref{eq:am} to obtain altered trajectories\;

        Compute $\bigtriangleup_l(s_t)$ and penalized advantage $A^{\lambda}(s_t,a_t)$ according to Eqs.~\eqref{eq:adv} and \eqref{eq:reg} \;
        
        Update the policy by maximizing the clipped surrogate function $J^{\mathit{CLIP}}$ according to Eq.~\eqref{eq:ppo}\;
        
        Update the value function network by minimizing the squared-error loss $L^V$ according to Eq.~\eqref{eq:loss2}\;
	}  
\end{algorithm}
\DecMargin{1em}

\section{Experiments}
For demonstrating the performance of our proposed method, we make use of the simulation environments \cite{d2020fairness,atwood2019fair} that implement toy examples of dynamic systems for supporting studies of long-term consequences of ML-based decision systems. We conduct three case studies in the context of bank loans, allocation of attention, and epidemic control. The proposed method is evaluated with respect to utility, short-term fairness, and long-term fairness. 
As mentioned earlier, the policies will be first trained by interacting with the environment and then tested separately in the environment. All the code and data are available online at \href{https://github.com/yaoweihu/Fairness-in-RL}{https://github.com/yaoweihu/Fairness-in-RL}.


{\bf\noindent Baselines}.
We consider two different categories of baseline agents in our experiments. The first category is human-designed policy agents used in \cite{d2020fairness,yu2022policy}, including the EO agent for bank loans, the CPO agent for attention allocation, and the Max agent for epidemic control.
The second category consists of learning-based policy agents. We consider the original PPO algorithm that only maximizes the cumulative reward, and the A-PPO algorithm that is the state-of-the-art fair RL algorithm proposed in \cite{yu2022policy} for achieving fairness through advantage regularization. As an ablation study, we also consider a variant of our method named F-PPO-L that only consists of the long-term fairness component but with the short-term fairness component removed.



\subsection{Case Study: Bank Loans}
In this case study, the bank lending scenario is simulated where an agent plays the role of a bank to make decisions about whether to grant loans to a stream of applicants. The qualification of applicants is described by a discrete credit score, which changes with the loan decisions.


\begin{figure*}[t]
    \centering
    \begin{subfigure}{0.32\textwidth}
        \includegraphics[width=\textwidth]{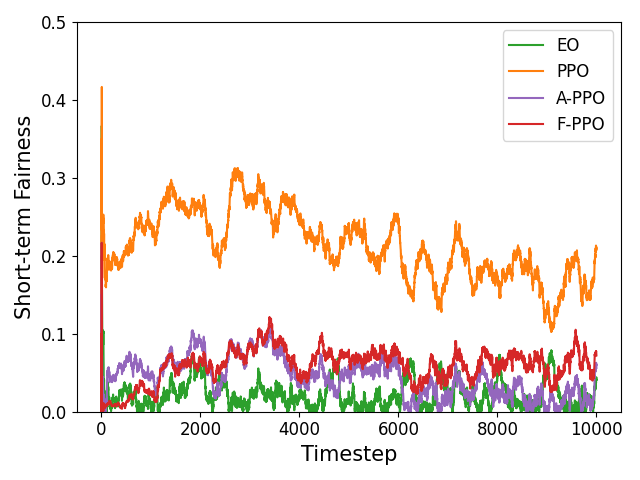}
        \caption{The short-term fairness}
        \label{fig:lendingsub1}
    \end{subfigure}
    \begin{subfigure}{0.32\textwidth}
        \includegraphics[width=\textwidth]{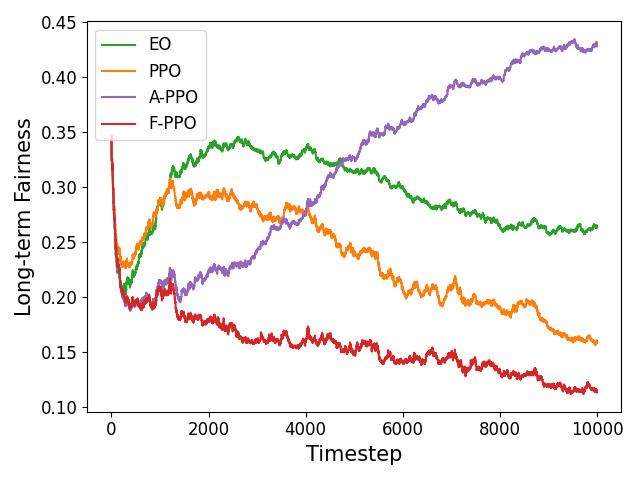}
        \caption{The long-term fairness}
        \label{fig:lendingsub2}
    \end{subfigure}
    \begin{subfigure}{0.32\textwidth}
        \includegraphics[width=\textwidth]{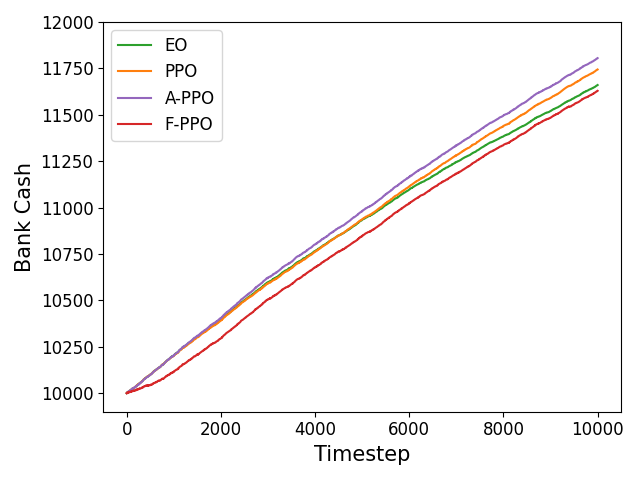}
        \caption{The amount of bank cash}
        \label{fig:lendingsub3}
    \end{subfigure}
    \caption{Experimental results for bank loans. The recorded values are averages over 10 evaluation runs.}
    \label{fig:lending}
\end{figure*}

{\bf\noindent Environment}.
In this environment, at each time step the bank observes a loan applicant $s_t$ which is sampled with replacement from the pool of applicants. Each applicant consists of a credit score (qualification feature) and a group membership (sensitive feature). The group membership $c_t$ is uniformly sampled from $\{c^+, c^-\}$. The credit score $x_t$, on the other hand, is drawn from a group-dependent discrete distribution over $X\in \{1,2,..., X_{max}\}$ with $\mathbb{E}_{c^-}[X] < \mathbb{E}_{c^+}[X]$. The bank employs a policy to make binary decisions of whether to deny or approve loan applications. If a loan applicant receives a loan and defaults, his/her credit score $C_i$ drops, which is simulated in the distribution by moving a small portion of mass from $P_{c}(X_i)$ to $P_{c}(X_{i-1})$ if $i\neq 1$. Similarly, if the loan applicant receives a loan and repays, a small portion of mass will be moved from $P_{c}(X_i)$ to $P_{c}(X_{i+1})$ if $i\neq X_{max}$. There is no change in the distribution if the applicant does not receive the loan. The bank's profit increases by the amount of the loan plus the interest on successfully repaid loans, and decreases by the loan amount on defaults. The probability of default is given by a deterministic function of the credit score. The reward of the bank at each time step is defined as the change in its profit at the next time step.


{\bf\noindent Implementation of F-PPO}.
In this case study, we implement a policy network $\pi_{\theta}(a_t|s_t)$ to make binary decisions $a_t\in \{1,0\}$. We adopt EO as the short-term fairness notion, which is defined as follows:
\begin{equation}
\begin{split}
&\bigtriangleup_s(s_t, a_t) = \\
&\left|\frac{\sum_{t-w}^{t} \text{successful\_loan}_{tc^+}}{\sum_{t-w}^{t}\text{will\_repay}_{tc^+}} - \frac{\sum_{t-w}^{t} \text{successful\_loan}_{tc^-}}{\sum_{t-w}^{t}\text{will\_repay}_{tc^-}}\right|,
\end{split}
\end{equation}
where $w = 300$ is the sliding window size. 
For the action massaging, the action will be flipped if the alternative action is fairer in terms of short-term fairness and the confidence of the action is lower than the threshold.
The threshold in Eq.~\eqref{eq:am} is dynamically adjusted according to the number of training iterations. 
The idea is to perform a cold start in action massaging so that the actions are not altered at the beginning of training. The threshold is initially zero and increases after a certain number of iterations.
Specifically, the threshold $\tau$ at the $i$th iteration is defined as $\tau=1-2\tau(i)$ where
\begin{equation*}
    \tau(i) = 
\begin{cases}
\tau_s \cdot \gamma^{i - i_s} & i \ge i_s \\
0.5 & \text{otherwise}
\end{cases}
\end{equation*}
In our experiments, we set $\tau_s = 0.5$, $i_s = 17$, and $\gamma = 0.985$. 
Finally, long-term fairness is computed as
\begin{equation*}
    \bigtriangleup_{l}(s_t)=W(P_{t-w:t}(X|c^+),P_{t-w:t}(X|c^-))
\end{equation*}
where $P_{t-w:t}(X|c)$ is the distribution mass of the credit score of group $c$ measured within the sliding window. For other hyperparameters, we set $\lambda = 1$ and $\delta = 0.05$.


{\bf\noindent Agents}. We include PPO, A-PPO, and EO as baselines to compare with our F-PPO, where EO is the agent that maximizes the bank profits subject to constraints of equal opportunity at every time step.

{\bf\noindent Results}. 
The short-term fairness, long-term fairness, and reward obtained by different agents during the test are shown in Fig. \ref{fig:lending}. As can be seen, despite similar performance in terms of reward (Fig. \ref{fig:lendingsub3}), the fairness performance of different methods is diverse. For short-term fairness (Fig. \ref{fig:lendingsub1}), EO, A-PPO and F-PPO are all able to keep the bias values below 0.1, while the original PPO produces much larger bias values. For long-term fairness (Fig. \ref{fig:lendingsub2}), our F-PPO approach exhibits superior performance among all the methods considered, as it consistently achieves the smallest bias values, and these bias values continue to decrease over time. On the other hand, A-PPO produces the worst performance where the bias values increase over time. This result shows that simply adding traditional fairness constraints into a long-term objective does not necessarily achieve long-term fairness. By combining the three results, we see that our F-PPO algorithm strikes a desirable balance between short-term fairness, long-term fairness, and utility of the policy.

\begin{figure*}[t]
    \centering
    \begin{subfigure}{0.32\textwidth}
        \includegraphics[width=\textwidth]{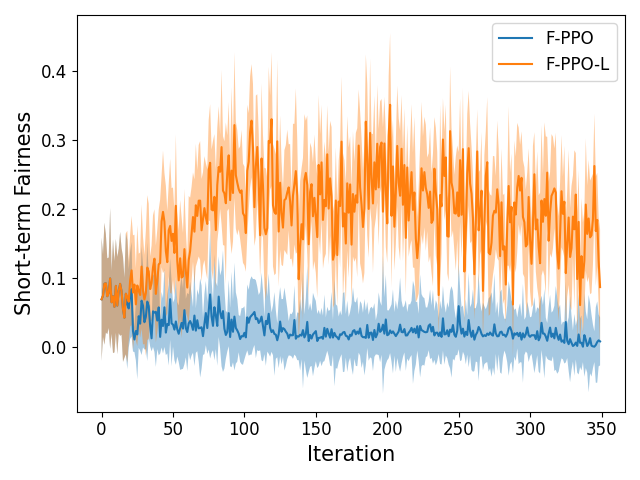}
        \caption{Bank Loans}
        \label{fig:assub1}
    \end{subfigure}
    \begin{subfigure}{0.32\textwidth}
        \includegraphics[width=\textwidth]{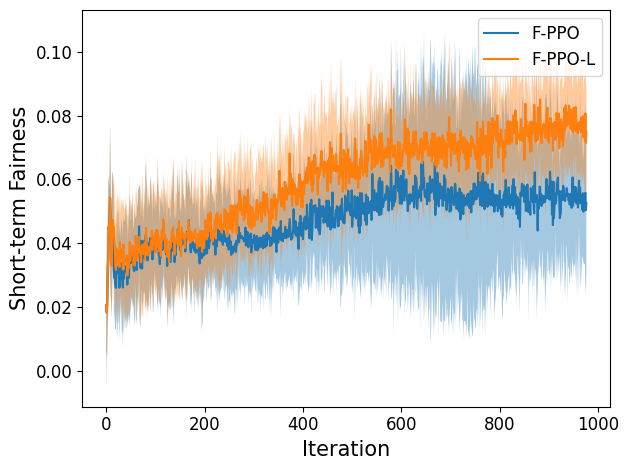}
        \caption{Attention Allocation}
        \label{fig:assub2}
    \end{subfigure}    
    \begin{subfigure}{0.32\textwidth}
        \includegraphics[width=\textwidth]{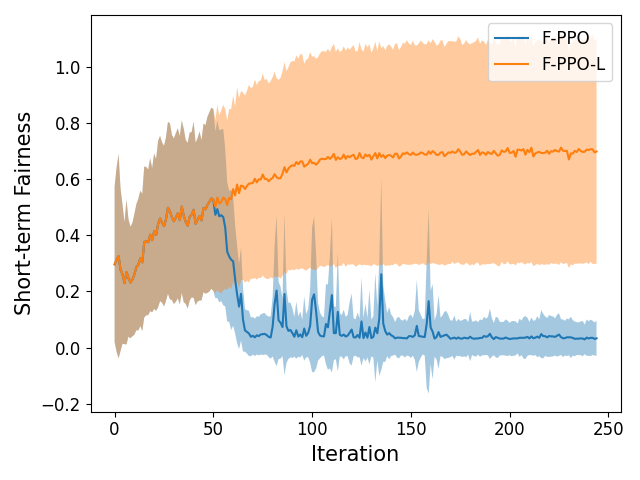}
        \caption{Epidemic Control}
        \label{fig:assub3}
    \end{subfigure}
    \caption{Ablation study: mean and standard deviation of short-term fairness in each iteration measured during training.}
    \label{fig:as}
\end{figure*}

For the ablation study, Fig. \ref{fig:assub1} shows the mean and standard deviation of short-term fairness achieved by F-PPO and F-PPO-L in each iteration of the training process over 350 iterations. The results show the effectiveness of the action massaging in maintaining short-term fairness throughout both the training and
evaluation processes, while the long-term fairness component alone cannot guarantee short-term fairness.

\subsection{Case Study: Attention Allocation}
This scenario aims to simulate incident monitoring and mitigation. In the simulation, the agent's role is to assign attention units to a set of locations. Each attention unit can prevent, or catch, one incident at the location to which it is assigned. The incident rates at each location vary over time in accordance with the number of incident occurrences as well as the agent's decisions on how to assign attention units.

\begin{figure*}[t]
    \centering
    \begin{subfigure}{0.32\textwidth}
        \includegraphics[width=\textwidth]{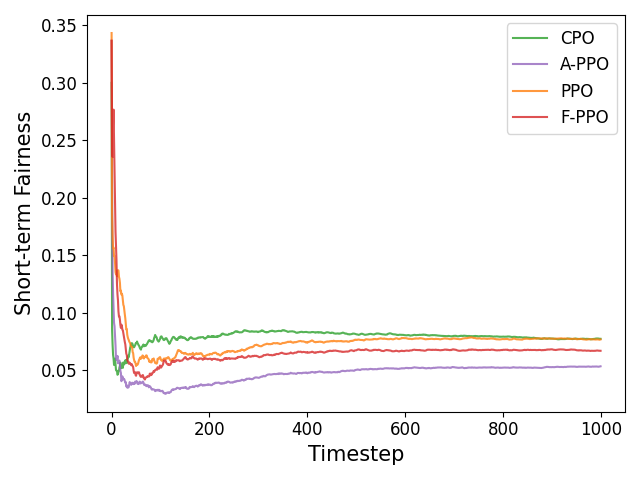}
        \caption{Short-term fairness}
        \label{fig:attnsub1}
    \end{subfigure}
    \begin{subfigure}{0.32\textwidth}
        \includegraphics[width=\textwidth]{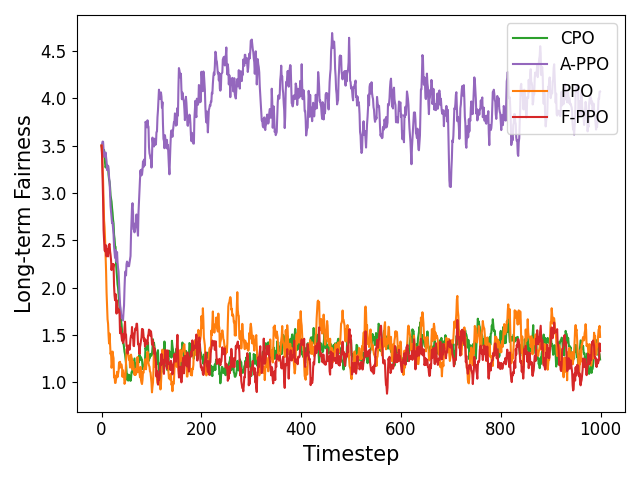}
        \caption{Long-term fairness}
        \label{fig:attnsub2}
    \end{subfigure}
    \begin{subfigure}{0.32\textwidth}
        \includegraphics[width=\textwidth]{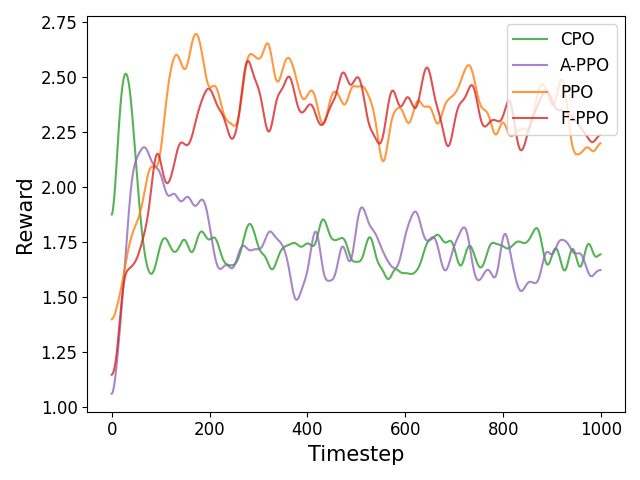}
        \caption{Average rewards}
        \label{fig:attnsub3}
    \end{subfigure}
    \caption{Results for the Attention Allocation environment. The recorded values are the averages over 10 evaluation episodes.}
    \label{fig:attn}
\end{figure*}

{\bf\noindent Environment}.
In the attention allocation environment, let $N$ represent the number of attention units, and $K$ be the number of locations. At each time step $t$, the agent assigns all $N$ units over the $K$ locations and $a_{k,t}$ denotes the number of units assigned to location $k$. The number of incidents that occur at each location is sampled from a Poisson distribution as $y_{k,t}\sim \text{Poisson}(\Lambda_{k,t})$, where $\Lambda_{k,t}$ is a dynamic parameter which changes according to
\begin{equation*}
    \Lambda_{k,t+1} = 
\begin{cases} \Lambda_{k,t}+\lambda_{k}^I & \text{if } a_{k,t}=0 \\
\Lambda_{k,t}-\lambda^D \cdot a_{k,t} & \text{otherwise.}
\end{cases}
\end{equation*}
Here, $\lambda_{k}^I$ is the increase rate, which may vary between locations $k$, and $\lambda_D$ is the decrease rate which is the same accross locations. 
The number of incidents discovered at a location is given by $\hat{y}_{k,t}=\min(a_{k,t},y_{k,t})$. The reward is defined as $r(s_t)=\zeta_0 \sum_{k=1}^{K}\hat{y}_{k,t}-\zeta_1\sum_{k=1}^{K}(y_{k,t}-\hat{y}_{k,t})$ which is determined by the fraction of incidents discovered. The parameters $\zeta_0$ and $\zeta_1$ weight the reward function in terms of incidents discovered versus incidents missed. The state $s_t$ is an observation history of length $H$ and each observation is a tuple of vectors $(\hat{y}_t,y_t,a_t,\hat{y}_t\oslash y_t)$, where $\oslash$ denotes the Hadamard division operation.

{\bf\noindent Implementation of F-PPO}. 
The policy network for attention allocation produces a $K$ dimensional vector of logits which are converted into a probability distribution $P(k)$ using the softmax function. The action is constructed by iteratively assigning attention units to the locations until all have been assigned. In each iteration, one attention unit is assigned to the location with the highest probability, from which the amount of $\frac{1}{N}$ is removed before the next iteration. For short-term fairness, we adopt DP as the metric defined as follows:
\begin{equation}
        \bigtriangleup_s(s_t, a_t) = \max_{k}\left|\frac{\sum_{t'=t-w}^{t} a_{k,t'}}{N\cdot w} - \frac{1}{K} \right|,
\end{equation}
which requires that the number of units assigned should be equal across different locations. The action massaging checks for each pair of locations $k_{1},k_{2}$ where at least one unit is assigned to $k_1$, if reallocating one unit from $k_{1}$ to $k_{2}$ would improve short-term fairness. To minimize the impact of the action massaging on the utility, the algorithm also checks if the difference between $P(k_1)$ and $P(k_2)$ is less than the threshold. When both conditions are met by multiple pairs, the algorithm selects the one that leads to the best short-term fairness performance and performs the reallocation. For simplicity, we use a static threshold of 0.08. Finally, we measure long-term fairness according to the incident distribution over all locations. In training, we estimate the incident distribution based on the number of incident occurrences, but in evaluation, we use $\Lambda_{k,t}$ as the ground truth of the incident distribution.

\begin{figure*}[t]
    \centering
    \begin{subfigure}{0.32\textwidth}
        \includegraphics[width=\textwidth]{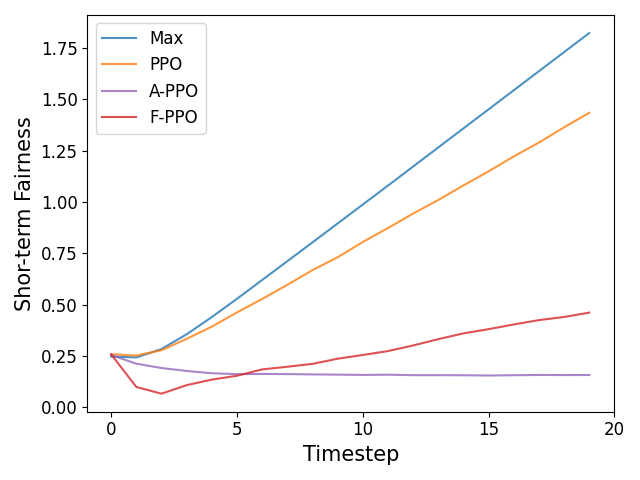}
        \caption{Short-term fairness}
        \label{fig:infsub2}
    \end{subfigure}
    \begin{subfigure}{0.32\textwidth}
        \includegraphics[width=\textwidth]{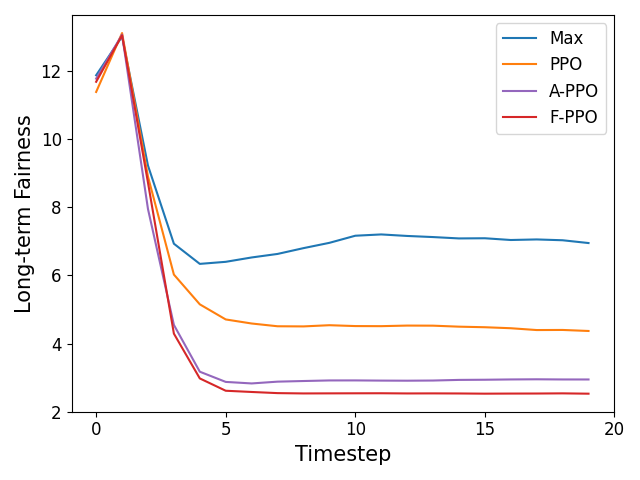}
        \caption{Long-term fairness}
        \label{fig:infsub3}
    \end{subfigure}
    \begin{subfigure}{0.32\textwidth}
        \includegraphics[width=\textwidth]{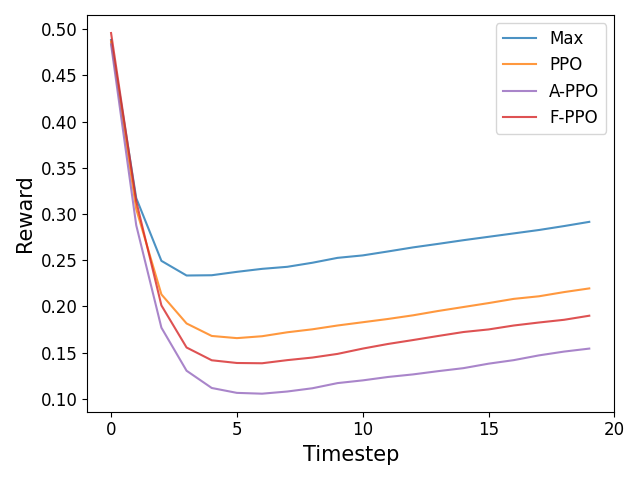}
        \caption{Average rewards}
        \label{fig:infsub1}
    \end{subfigure}
    
    \caption{Experimental results for epidemic control. The recorded values are averages over 200 evaluation episodes.}
    \label{fig:inf}
\end{figure*}

{\bf\noindent Agents}. For baselines used to evaluate our F-PPO agent, we consider PPO, A-PPO, as well as the CPO agent that aims to discover the most incidents.

{\bf\noindent Results}. The experimental results are shown in Fig. \ref{fig:attn}. Our F-PPO achieves the best long-term fairness performance while maintaining relatively low short-term bias values and high rewards. As a comparison, although A-PPO produces the best short-term fairness performance, its long-term fairness performance and utility are among the worst. For the ablation study, Fig. \ref{fig:assub2} shows the effectiveness of the short-term fairness component.

\subsection{Case Study: Epidemic Control}
The third case study simulates an infectious disease scenario where vaccines are allocated within a social network in a step-by-step manner.
At each step, one individual is selected by the policy to receive the vaccine. Meanwhile, healthy individuals have the chance to get infected, and sick individuals have the chance to recover. The task is to optimally allocate vaccines in order to mitigate the spread of disease.


{\bf\noindent Environment}. 
In this environment, we have a social network $G=(V,E)$ where $V$ is a set of individuals and $E$ is a set of edges representing social connections. The state of the environment is a vector of the health state of all individuals. For each individual, the health state is represented by a four-dimensional one-hot encoding $H=\{S,I,R\}$ that represents the three possible states that an individual can be in, including susceptible (healthy), infected, and recovered. At the initial time step, a random set of individuals $V_0$ are infected. Then, at each time step $t$, the probability of a susceptible individual $i$ transitioning to the state of infected is given by $P_{I}(v_{i,t})= 1-(1-\tau)^{|N_{I}(v_{i,t})|}$, where $N_{I}(v_{i,t})$ represents the number of infected individuals in the neighbor of individual $i$ and $\tau \in [0,1]$ is an infectious factor. Meanwhile, the probability of an infected individual recovering is given by $P_{R}(v)=\rho$ where $\rho\in [0,1]$ is the recovering factor. If a susceptible individual receives the vaccine, his/her state directly transitions from S to R. The reward is defined as the proportion of the population who are not infected. To study fairness, the Girvan-Newman algorithm \cite{girvan2002community} is used to partition the network $G$ into two communities corresponding to groups $c^+$ and $c^-$.


{\bf\noindent Implementation of F-PPO}. 
The policy network is a multiclass classifier that outputs the probabilities of $|V|+1$ actions representing either not vaccinating or vaccinating any of the $|V|$ individuals. EO is still adopted as the short-term fairness metric, which measures the vaccination ratio among newly infected individuals in different groups as follows
\begin{equation}
    \begin{split}
        &\bigtriangleup_s(s_t, a_t) = \\ 
        &\left|\frac{\sum_{t-w}^{t} \text{vaccine\_given}_{tc^+}}{\sum_{t-w}^{t}\text{new\_infected}_{tc^+}+1} - \frac{\sum_{t-w}^{t} \text{vaccine\_given}_{tc^-}}{\sum_{t-w}^{t}\text{new\_infected}_{tc^-} + 1}\right|.
    \end{split}
\end{equation}
For the action massaging, at each time step it checks if providing the vaccine to an individual from the other community would result in a fairer allocation. If this condition is met, the algorithm proceeds to check if there exists an individual from the other community whose predicted probability is sufficiently close to that of the current individual. If such an individual is found, the algorithm modifies the action accordingly. A dynamic threshold is again adopted. For the $i$th iteration, the threshold is defined as:
\begin{equation*}
    \tau(i) = 
\begin{cases} \min(\tau_e, \tau_s \cdot \gamma^{i - i_s}) & i \ge i_s \\
0 & \text{otherwise}
\end{cases}
\end{equation*}
Specifically, we set $\tau_s = 0.01$, $\tau_e = 0.35$, $i_s = 50$, $\gamma = 1.2$. Finally, long-term fairness is measured as the distance between the health states of the two communities. For other hyperparameters, we set $\lambda = 0.25$ and $\delta = 0.05$.

{\bf\noindent Agents}. We consider the Max agent in addition to PPO, A-PPO, and F-PPO. The Max agent vaccinates the most susceptible individual each time, which is considered as the individual with the most number of infected neighbors.


{\bf\noindent Results}.
The experimental results are shown in Fig. \ref{fig:inf}. As can be seen, F-PPO achieves the best performance in terms of long-term fairness and significantly improves short-term fairness compared with the Max and PPO agents. A-PPO achieves the best performance in terms of short-term fairness, but produces the worst performance in gaining rewards. As expected, the Max agent achieves the highest utility performance, but it also demonstrates the poorest fairness performance. The combination of the results also demonstrates the capability of the F-PPO. The ablation study in Fig. \ref{fig:assub3} shows similar results to the other two case studies.




\section{Conclusions}
In this paper, we studied the problem of achieving long-term fairness in sequential decision-making systems. We modeled the system as a Markov Decision Process (MDP) and tackled the problem by developing a fair reinforcement learning (RL) algorithm. By acknowledging
that short-term fairness and long-term fairness are distinct requirements that may not necessarily align with one another, we developed an algorithmic framework that incorporates both requirements using different bias mitigation approaches, including pre-processing and in-processing approaches. We conducted three simulation case studies. The results show that our method can strike a balance between
short-term fairness, long-term fairness, and utility.


\section*{Acknowledgments}
This work was supported in part by NSF 2142725.

\bibliographystyle{IEEEtran}
\bibliography{main}

\end{document}